\documentclass[10pt,twocolumn,letterpaper]{article}

\usepackage{3dv}

\makeatletter
\@namedef{ver@everyshi.sty}{}
\makeatother
\usepackage{tikz}

\usepackage{times}
\usepackage{epsfig}
\usepackage{graphicx}
\usepackage{amsmath}
\usepackage{amssymb}

\usepackage[ruled]{algorithm2e} 

\SetAlFnt{\small}
\SetAlCapFnt{\small}
\SetAlCapNameFnt{\small}
\SetAlCapHSkip{0pt}

\graphicspath{{figures/}}
\usepackage{bbold}

\usepackage{amsfonts}
\usepackage{amsmath}
\usepackage{amssymb}
\usepackage{overpic}
\usepackage{pgfplots}
\usepackage{pgf,tikz}
\usepackage{bm}
\usepackage{wrapfig}
\usepackage{multirow}
\usepackage{mathtools}
\usepackage{enumerate}
\usepackage{verbatim}

\usepackage{tabls}

\renewcommand{\phi}{\varphi}

\newcommand{\RR}{\mathbb{R}}

\newcommand{\Ee}{\mathcal{E}}

\newcommand{\Ss}{\mathcal{S}}

\newcommand{\al}{\alpha}

\newcommand{\lb}{\lambda}

\def\*#1{\mathbf{#1}}

\DeclareMathOperator*{\argmin}{arg\,min}

\usepackage{xcolor}
\definecolor{tabblue}{rgb}{0.54, 0.81, 0.94}

\usepackage{listings}
\usepackage{color, colortbl}
\definecolor{mygreen}{RGB}{28,172,0} 
\definecolor{mylilas}{RGB}{170,55,241}
\definecolor{myblue}{RGB}{203,192,211}
\definecolor{mycolor1}{RGB}{255,204,201}

\usepackage{booktabs}       
\newlength{\Oldarrayrulewidth}

\usepackage{amssymb}

\usepackage[shortlabels]{enumitem}
\usepackage{kantlipsum}

\newcommand*{\Scale}[2][4]{\scalebox{#1}{$#2$}}%
\usepackage[hang,flushmargin]{footmisc}

\usepackage[pagebackref=true,breaklinks=true,colorlinks,bookmarks=false]{hyperref}

\newcommand{\mypara}[1]{\vspace{-3mm} \paragraph{#1}}

\threedvfinalcopy 

\def\threedvPaperID{332} 

\ifthreedvfinal\fi
\begin{document}

\title{Smooth Non-Rigid Shape Matching via Effective Dirichlet Energy Optimization}

\author{Robin Magnet\\
LIX, \'Ecole Polytechnique, IP Paris \\
{\tt\small rmagnet@lix.polytechnique.fr}
\and
Jing Ren\\
ETH Zurich\\
{\tt\small jing.ren@inf.ethz.ch}
\and
Olga Sorkine-Hornung\\
ETH Zurich \\
{\tt\small sorkine@inf.ethz.ch}
\and
Maks Ovsjanikov\\
LIX, \'Ecole Polytechnique, IP Paris \\
{\tt\small maks@lix.polytechnique.fr}
}

\maketitle


\begin{abstract}
We introduce pointwise map smoothness via the Dirichlet energy into the functional map pipeline, and propose an algorithm for optimizing it efficiently, which leads to high-quality results in challenging settings. Specifically, we first formulate the Dirichlet energy of the pulled-back shape coordinates, as a way to evaluate smoothness of a pointwise map across discrete surfaces. We then extend the recently proposed discrete solver and show how a strategy based on auxiliary variable reformulation allows us to optimize pointwise map smoothness alongside desirable functional map properties such as bijectivity. This leads to an efficient map refinement strategy that simultaneously improves functional and point-to-point correspondences, obtaining smooth maps even on non-isometric shape pairs. Moreover, we demonstrate that several previously proposed methods for computing smooth maps can be reformulated as variants of our approach, which allows us to compare different formulations in a consistent framework. Finally, we compare these methods both on existing benchmarks and on a new rich dataset that we introduce, which contains non-rigid, non-isometric shape pairs with inter-category and cross-category correspondences. Our work leads to a general framework for  optimizing and analyzing map smoothness both conceptually and in challenging practical settings.
\end{abstract}

\section{Introduction}
Shape correspondence is a fundamental task in Geometry Processing, acting as a building block for many downstream applications \cite{van2011survey,sahilliouglu2020recent,deng2022survey}. One of the key challenges in designing a successful general-purpose shape matching method is the choice of the objective function that should promote high quality correspondences and, at the same time, be easy enough to optimize in order to be applicable on complex, densely sampled geometric objects.

A widely acknowledged desirable objective in non-rigid shape matching is \textit{smoothness}, which intuitively promotes local consistency or continuity of computed correspondences, while being less restrictive than, e.g., isometries or conformal maps. Several works have incorporated smoothness into the map computation pipelines either via auxiliary energy terms  \cite{RHM}, or by structuring the search space privileging continuous, often low frequency, correspondences or deformation fields, e.g., \cite{eisenberger2019divergence,smoothshells}. Despite the utility of smoothness as a supervising signal in map computation, existing approaches can either be difficult to scale to dense meshes or are incorporated in an ad-hoc manner. Moreover, there is no coherent framework for comparing different existing strategies for promoting map smoothness using a single consistent computational and conceptual formalism.

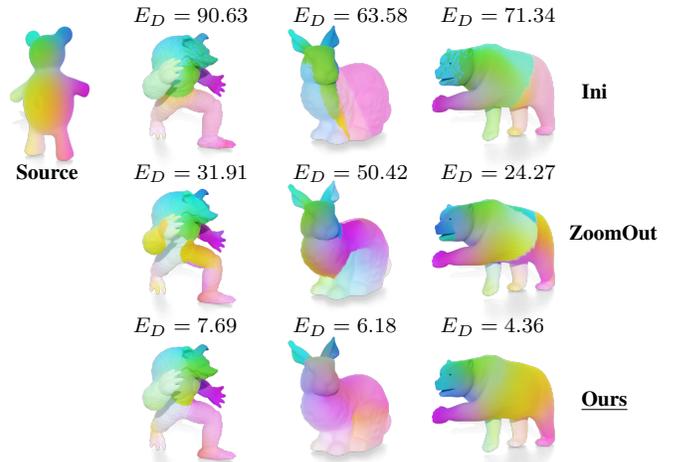
\begin{figure}[!t]
\centering
 \begin{overpic}[trim=0.8cm 0.4cm 0.2cm 0.85cm,clip,width=0.98\linewidth,grid=false]{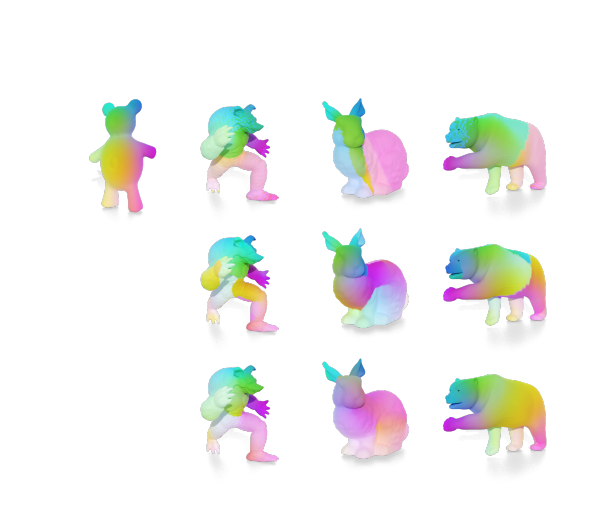}
 \put(3,48){\footnotesize\bfseries Source}
 \put(95,61){\footnotesize\bfseries Ini}
 \put(93,38){\footnotesize\bfseries ZoomOut}
 \put(95,11){\footnotesize\bfseries \underline{Ours}}
 
 \put(22,73.5){\footnotesize $E_D = 90.63$}
 \put(48,73.5){\footnotesize $E_D = 63.58$}
 \put(72,73.5){\footnotesize $E_D = 71.34$}
 
 \put(22,48){\footnotesize $E_D = 31.91$}
 \put(48,48){\footnotesize $E_D = 50.42$}
 \put(72,48){\footnotesize $E_D = 24.27$}

 \put(22,23){\footnotesize $E_D = 7.69$}
 \put(48,23){\footnotesize $E_D = 6.18$}
 \put(72,23){\footnotesize $E_D = 4.36$}

\end{overpic}\vspace{-6pt}
\caption{Our method can deal with noisy inputs and produce high-quality and smooth pointwise maps for non-isometric shape pairs. As a comparison, ZoomOut~\cite{zoomout}, the current state-of-the-art refinement method, cannot explicitly control  the map smoothness and can have large discontinuous patches in the obtained maps. We report the smoothness metric $E_D$ for each map.}
\label{fig:teaser}\vspace{-12pt}
\end{figure}

\begin{figure*}[!t]
    \centering\vspace{-6pt}
    \begin{overpic}[trim=0cm 0cm 0cm 6.8cm,clip,width=1\linewidth,grid=false]{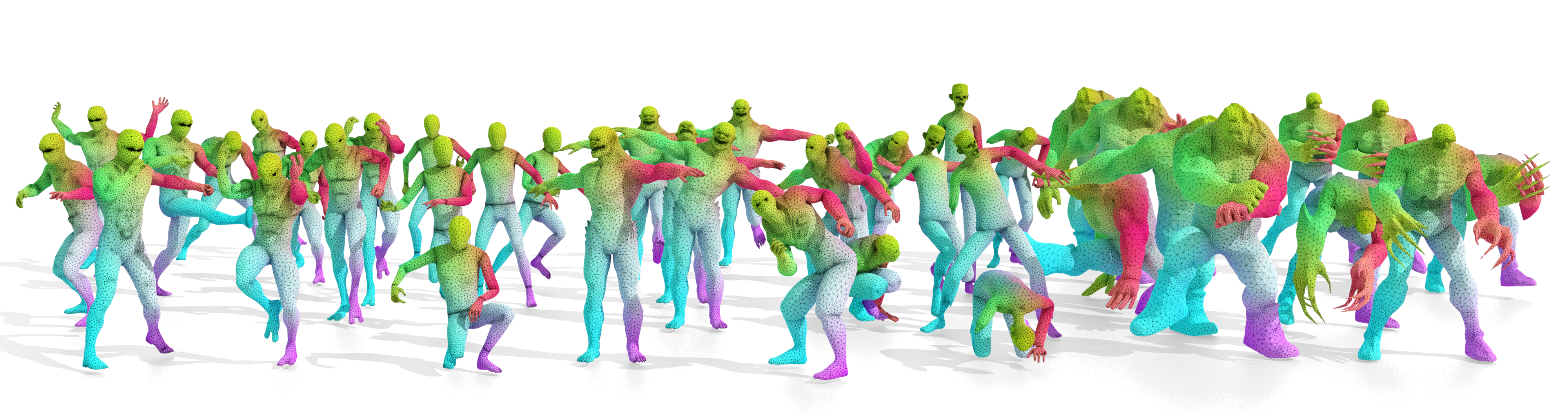}
    \end{overpic}
    \vspace{-1cm}
    \caption{\textbf{\textsc{DeformThings4D-Matching} Dataset.} We construct a new dataset for non-isometric shape matching based on the \textsc{DeformThings4D}~\cite{li20214dcomplete}. We show some example humanoid shapes and visualize the cross/inter-category correspondences via color transfer. Note that the shapes in the same category are remeshed independently (zoom in to see the mesh wireframes).}
    \label{fig:res:dt_deform4d}\vspace{-12pt}
\end{figure*}


In this paper we focus on the functional map framework, which was originally proposed as a tool for near-isometric shape matching \cite{ovsjanikov2012functional} and has since then been significantly extended to different tasks \cite{rustamov2013map,huang2019limit} and correspondence models \cite{rodola2017partial,kovnatsky2013coupled,donati2020deep}, among many others.
The key advantages of this framework are its efficiency and flexibility. The efficiency of functional maps-based approaches stems from representing maps as small matrices using a reduced basis, which leads to small-scale optimization problems. At the same time, this framework is flexible and can incorporate a wide range of desirable constraints using simple linear algebraic formulations, e.g., \cite{eynard2016coupled,nogneng17,ovsjanikov2017computing}.

Although originally functional map-based methods focused on constraints in the functional (spectral) domain, recent works have started to highlight and exploit links that exist between pointwise and functional map representations, while leveraging the strengths of both \cite{ren2018continuous,RHM,ren2021discrete}. Specifically, a recent discrete optimization scheme was proposed in \cite{ren2021discrete}, demonstrating that many desirable map properties can be optimized directly in the pointwise map representation. Unfortunately, while the class of energies considered in \cite{ren2021discrete} covers many existing functional map objectives, such as bijectivity or commutativity with the Laplacian, it does not address desirable pointwise map properties such as map smoothness. This can lead to local inconsistencies, such as discontinuous mapped patches, thus severely limiting the utility of the computed maps in practice.

In this paper we introduce a novel method that allows to explicitly promote pointwise map smoothness within the functional map framework. Our method is based on, first, formulating smoothness as the optimization of the Dirichlet energy of the pointwise map, and second, an iterative method for solving this energy optimization by extending the method introduced in \cite{ren2021discrete}. This allows our approach to be used alongside other desirable objectives, while explicitly promoting smooth and locally consistent maps. We therefore both extend the scope of discrete map optimization to new energies not covered in \cite{ren2021discrete} and use this insight to develop an efficient non-rigid shape matching approach that directly promotes pointwise map smoothness.

In addition to introducing a novel method for promoting smooth maps within the functional maps framework, we also investigate multiple previous approaches for computing smooth maps in different settings \cite{RHM,smoothshells,arap,nicp} and show how they can be interpreted as variants of each other and thus compared within a unified formalism. This allows us to design a \textit{family} of different approaches, parametrized by the choice of the smoothness energy and its associated optimization strategy. We propose a coherent formalism within which various energies can be compared and demonstrate their relative utility in different settings. %
Finally, we observe that most public datasets focus on near-isometric pairs, making it non-trivial to evaluate accuracy and smoothness in more realistic scenarios, which can involve diverse and non-isometric shapes. To fill this gap we introduce a new challenging dataset based on \textsc{DeformThings4D}~\cite{li20214dcomplete}, but with additional cross-category ground truth maps (Fig.\ \ref{fig:res:dt_deform4d}). We use this dataset alongside existing benchmarks in a comprehensive comparison of various approaches computing smooth correspondences.
To summarize, our key \textbf{contributions} include:
\begin{enumerate}[leftmargin=*, noitemsep]
	\item We show how pointwise map smoothness can be formulated and optimized within the functional map framework, by extending the discrete solver proposed in \cite{ren2021discrete}.
	\item Based on this construction, we introduce a simple and effective map refinement method that is both computationally efficient and leads to high-quality results in non-isometric settings (Fig.~\ref{fig:teaser}).
	\item We show how several previously proposed methods are intimately related both to our approach and within themselves, and propose a coherent framework, allowing us to directly compare ways to promote smoothness within a consistent formalism and computational strategy.
	\item We construct a new dataset for non-rigid shape matching tasks with inter-category correspondences for animal shapes, and inter-/cross-category correspondences for humanoid shapes that are independently remeshed.
\end{enumerate}

\section{Related Work}\label{sec:related_work}

In this section, we briefly review the previous works of shape matching, commonly used map evaluation metrics, and various map solvers, that are most related to this work. We refer to recent surveys~\cite{deng2022survey,sahilliouglu2020recent,biasotti2016recent} for more thorough discussions of shape matching.

\mypara{Shape Matching}
Our work focuses on the problem of shape matching, that looks for dense correspondences between two non-rigid 3D shapes.
One solution to shape matching is to solve for correspondences directly by minimizing an explicit and carefully designed energy~\cite{bronstein2006,huang2008,ovsjanikov2010}, which can lead to complex combinatorial problems with high computational complexity.
An alternative solution is to find correspondences between parametric representations, where the input shapes are mapped into a canonical domain~\cite{Lipman2009,Aigerman15,tutte}.
Our work is based on the functional map representation~\cite{ovsjanikov2012functional,ovsjanikov2017computing}, which computes correspondences between functions defined on the shapes. %
Different regularizers have been proposed to promote the accuracy of functional maps~\cite{nogneng17,nogneng18,ren2018continuous,gehre2018interactive,wang2018kernel,wang2018vector}.
Computing a functional map is usually reduced to solving a least-square system, which has a relatively low computation cost, but recovering a point-wise correspondence from the computed functional map is error-prone~\cite{rodola-vmv15,ezuz2017deblurring,ren2021discrete}.
To further improve the accuracy of the recovered point-wise correspondences, different refinement methods have been proposed as a post-processing step~\cite{solomon2016entropic,mandad2017variance,vestner2017product,vestner2017efficient}. 
A common technique for map refinement in the functional maps framework is to iteratively update functional maps and the underlying pointwise maps according to different energies such as Dirichlet energy and bijectivity~\cite{ovsjanikov2012functional,RHM,ren2018continuous,zoomout,MapTree,ren2021discrete}.
In this work, we present a new refinement method that can robustly deal with noisy input and efficiently produce smooth maps in the functional maps framework.

\mypara{Metrics for Map Quality Evaluation}
Different criterion have been taken into consideration to evaluate map quality, which are incorporated into map computation.
The most commonly-used metric is the map accuracy, which is measured by comparing the geodesic distance between the mapped position and the pre-specified ground-truth position.
Some previous work~\cite{nicp,mandad2017variance,gehre2018interactive} adopt a landmark term to enforce map accuracy.
To achieve a fully automatic solution, other metrics such as smoothness, bijectivity, conformality, and coverage are considered for map optimization other than accuracy which needs manually specified landmarks.
For example, Reversible Harmonic Maps~\cite{RHM} proposes to optimize the Dirichlet energy together with the bijectivity of the pointwise maps.
Smooth Shells~\cite{smoothshells} adopts the ARAP energy~\cite{arap} to compute a smooth deformation field, which potentially leads to a smooth pointwise map.
\cite{kim2011blended}~blends across multiple maps to get a smooth one.
\cite{ren2018continuous} proposes heuristics to improve the bijectivity, smoothness, and coverage of the pointwise map in both spatial and spectral domain. 
In this work, we observe how several previous proposed approaches are closely related in formulating map smoothness. We show different variants can be compared in a coherent way within a consistent formalism.

\mypara{Map Solver}
Previous methods adopt different search space for maps and hence need different solvers. 
For example, some work~\cite{fogel2013convex,solomon2016entropic,dym2017ds++,sinkhorn} solve for maps that are represented by doubly stochastic matrices.
Functional maps framework~\cite{ovsjanikov2012functional,nogneng17,nogneng18,ren2018continuous} usually solves a least-square system for functional maps. 
Quadratic-splitting technique~\cite{ezuz2017deblurring,RHM} is also used to solve vertex-to-point (also called precise) maps.
\cite{ren2021discrete} introduces a discrete solver to optimize commonly used functional map energies constrained on the proper functional maps, which is a subset of functional maps that are associated with pointwise maps.
In this work, we introduce map smoothness into functional map pipeline and present an efficient algorithm to minimize the smoothness which extends the scope of discrete solver.

\section{Notation \& Background}\label{sec:background}

\mypara{Notation}
Given a triangle mesh $\Ss = (X, F)$ with the vertex positions $X$ and face set $F$, we denote the cotangent weight matrix by $W$ and the diagonal lumped mass matrix by $A$~\cite{meyer03}. By solving the generalized eigenvalue problem $W \phi_j = \lambda_j A \phi_j$, we can obtain the Laplace-Beltrami basis $\Phi$ by collecting the first $k$ eigenfunctions as columns, i.e., $\Phi = \big[ \phi_1 ... \phi_k\big]$ and the corresponding eigenvalues in a diagonal matrix, denoted as $\Delta = \text{diag}\big(\lambda_1 ... \lambda_k\big)$. We then have $\Phi^\top A \Phi = I$. 
A pointwise map is denoted as $\Pi_{ij}: \Ss_i \rightarrow \Ss_j$, where the subscript indicates the map direction. Specifically, $\Pi_{ij}\in \{0,1\}^{n_i\times n_j}$ ($n_i$ is the number of vertices in $\Ss_i$) is a binary matrix indicating the correspondences between the two shapes. For example, if the $p$-th vertex on $\Ss_i$ is mapped to the $q$-th vertex on shape $\Ss_j$, we then have $\Pi_{ij}(p,q) = 1$ and $\Pi_{ij}(p,t) = 0$ for $\forall t\neq q$.

\mypara{Functional Maps Framework}
The goal of shape matching is to find semantically meaningful and continuous pointwise map for a given shape pair. In this work, we follow the functional map framework~\cite{ovsjanikov2012functional} and encode a point-wise map as a linear transformation (called functional map) in the Laplace-Beltrami basis.
Specifically, for a pointwise map $\Pi_{ij}: \Ss_i \rightarrow \Ss_j$, the associated functional map is given as $C_{ji} = \Phi_i^{\dagger}\Pi_{ij}\Phi_j$. Note that $C_{ji}$ a pull-back linear operator that maps  functions on shape $\Ss_j$ to functions on shape $\Ss_i$.
In the original pipeline~\cite{ovsjanikov2012functional}, a functional map is computed by solving a least-squared system in the continuous linear operator space, i.e., $C_{21} = \argmin_{C\in \RR^{k_1\times k_2}} E(C)$, where $E(\cdot)$ is a functional map energy that preserves input descriptors or landmarks, surface area or angles, multiplicative operators, or shape orientation~\cite{ovsjanikov2012functional,nogneng18,ren2018continuous,huang2017adjoint}.
Solving for a function map in the unconstrained search space simplifies the optimization problem, but can lead to errors  when converting the computed functional map to a pointwise one~\cite{rodola-vmv15,ezuz2017deblurring,ren2021discrete}. Thus, additional post-processing techniques are used to improve the quality of the pointwise maps~\cite{ren2018continuous,RHM,zoomout,sinkhorn}.

\mypara{Discrete Optimization}
A recent work~\cite{ren2021discrete} has proposed a \emph{discrete solver} for functional map pipeline which constrains the optimization problem to the space of \emph{proper functional maps}.
Specifically  the functional map, 
$C_{21} = \argmin_{C\in \mathcal{P}_{21}} E(C)$, is solved in a discrete search space $\mathcal{P}_{21} = \big\{\,C_{21} \,\vert\, \exists \Pi_{12} \,\text{ s.t. }\, C_{21} = \Phi_1^{\dagger} \Pi_{12} \Phi_{2}\big\}$, i.e., the set of functional maps arising from \textit{some} pointwise correspondence.
The general strategy to solve this constrained problem, advocated in \cite{ren2021discrete} mimics the Augmented
Lagrangian methods with variable splitting~\cite{GABAY1976} and consists of the following two main steps: \textbf{(i)} reformulate the energy $E(\cdot)$ by making $C_{21}$ and $\Pi_{12}$ independent variables, and  adding a \emph{coupling} term:
\begin{equation}\label{eq:coupling}
    E_{\text{couple}}(C_{21}, \Pi_{12}) = \big\Vert C_{21} - \Phi_1^{\dagger} \Pi_{12} \Phi_{2}\big\Vert_F^2,
\end{equation}
\textbf{(ii)} iteratively solve for $C_{21}$ and $\Pi_{12}$ with the other variables fixed. This approach is shown to be efficient and leads to high-quality and well-regularized functional maps. 
Key to the success of this strategy is the ability to reformulate the given functional map energy so that the resulting optimization problems for $C_{21}$ and $\Pi_{12}$ in step (ii) can be solved in closed form. In~\cite{ren2021discrete}, a range of energies is considered including bijectivity, landmarks preservation, orthogonality and Laplacian commutativity. 
%

\mypara{Dirichlet Energy}
Given two Riemannian manifolds $\Ss_1$ and $\Ss_2$, the Dirichlet energy of a map $f: \Ss_1 \rightarrow \Ss_2$ is defined as $E_D(f) = \frac{1}{2}\int_{\Ss_1} \Vert df \Vert^2 d\mu_{\Ss_1}$, with $df$ the map differential, which intuitively acts as a measure of the stretch induced by the map (see, e.g., \cite{RHM} for a discussion). A \emph{smooth} map $f$ is therefore characterized as being a minimizer of the Dirichlet energy.
In the discrete setting, a map $f: \Ss_1 \rightarrow \Ss_2$ can be seen as a function between the two surface embeddings (i.e., $f:\RR^{3} \rightarrow \RR^{3}$) and is assumed to be affine on each face. We can then define the discrete Dirichlet energy~\cite{pinkall93}: 
\begin{equation}\label{eq:bg:dirichlet_def}
    E_D\left(f\right) = \sum_{(x_i, x_j)\in\Ee(\Ss_1)} w_{ij}\big\Vert f(x_i) - f(x_j)\big\Vert^2,
\end{equation}
where $\Ee(S_1)$ is the set of edges on $\Ss_1$ and $w_{ij}$ the cotangent weight of edge $(i,j)$.
We can rewrite Eq.~\eqref{eq:bg:dirichlet_def} in a more compact way:
    $E_D(f) = \text{Trace}(f^\top W_1 f) := \big\Vert f \big\Vert^2_{W_1}$,
where $W_1$ is the cotangent weight matrix of shape $\Ss_1$. 

Note that in practice one only needs to store the value of $f$ at each vertex of $\Ss_1$ and therefore if $f$ is a pointwise map from $\Ss_1$ to $\Ss_2$, we can represent it in matrix form $f=\Pi_{12}X_2$, where the value at row $i$ gives the coordinates $f(x_i)$. We therefore define the Dirichlet energy of the map $\Pi_{12}$ as the Dirichlet energy of $f$, which is the $W$-norm of the pull-back vertex coordinates:
\begin{equation}\label{eq:Dirichlet pull back}
    E_D\left(\Pi_{12}\right) = \big\Vert \Pi_{12}X_2\big\Vert^2_{W_1}.
\end{equation}
Note that \cite{RHM} adopts a similar formulation to measure the smoothness of a given map, but pulls-back a high-dimensional embedding, in which the $L^2$ distance approximates the geodesic distance, and that is computed via multidimensional scaling \cite{cox2008multidimensional}.



While the Dirichlet energy defines a measure of distortion induced by a map, we note that mapping all vertices in $\Ss_1$ to a single vertex in $\Ss_2$ leads to zero energy, as seen by setting $f(x_i)=y$ for some fixed $y$ in Eq.~\eqref{eq:bg:dirichlet_def}. The Dirichlet energy thus only contains partial information about the quality of the map, and one needs to use additional constraints to obtain a \textit{non-trivial} smooth map.

\section{Discrete Solver for Dirichlet Energy}\label{sec:mtd}
\begin{figure}[!t]
    \centering
    \begin{overpic}[trim=0.8cm 0.75cm 0.8cm 1.0cm,clip,width=1\linewidth,grid=false]{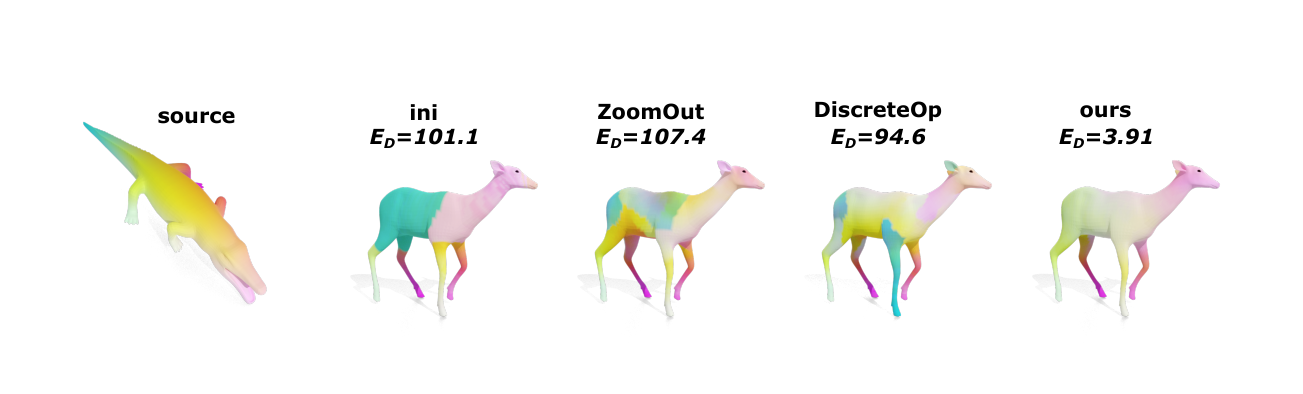}
    \end{overpic}\vspace{-6pt}
    \caption{Previous methods focus on improving map accuracy and do not have explicit control over the map smoothness. Here we show an example of a non-isometric pair. We report the Dirichlet energy ($E_D$) of maps after refinement by different methods.}
    \label{fig:mtd:eg_smooth}
\end{figure}

While functional maps intrinsically represent correspondences using low frequency eigenfunctions, thus inducing some smoothness, they do not provide any explicit control over the pointwise map smoothness (see Fig.~\ref{fig:mtd:eg_smooth}).
The discrete solver proposed in~\cite{ren2021discrete} has shown that many desirable map properties can be promoted directly on the functional maps, including bijectivity, landmarks preservation or conformality, the latter being unable to effectively promote smoothness as shown in the supplementary material.  In this work we therefore seek to extend this framework by introducing pointwise map smoothness constraint that can be efficiently used alongside other objectives.

\subsection{Problem Formulation}
As discussed in Sec.~\ref{sec:background}, the Dirichlet energy, seen as a measure of smoothness, is globally minimized by constant maps.
To avoid such trivial solutions, we propose to couple a smoothness energy with bijectivity constraints, which can be enforced in the spectral domain using the discrete optimization framework~\cite{ren2021discrete}.
%
%

Specifically, given two shapes $\Ss_1$ and $\Ss_2$ we consider functional maps $C_{ij}$ and pointwise maps $\Pi_{ij}$ from \emph{both} directions, where $(i,j) \in \{1,2\}^2$ indicates the map direction. 
%
%
The discrete solver framework~\cite{ren2021discrete} introduces a bijectivity energy which reads:
\begin{equation}\label{eq:mtd:E_biject}
   \Scale[0.92]{E_{_{\text{bij}}}\big( \Pi, C\big) = \sum\limits_{ij} \big\Vert \Pi_{ji}\Phi_i C_{ji} - \Phi_j \big\Vert_{A_j}^2 + \al \big\Vert \Phi_{j} C_{ij} - \Pi_{ji}\Phi_i \big\Vert_{A_j}^2},
\end{equation}
where the first term is derived from a spectral bijectivity energy and the second is a coupling term between functional maps $C_{ij}$ and pointwise maps $\Pi_{ji}$ (note the change in map directions). Note that variables $\Pi$ and $C$ contain maps in \emph{both} directions in order to simplify notations. We refer the reader to~\cite{ren2021discrete} for a detailed derivation.

In this work, we augment this energy using smoothness constraints, acting on the primal domain instead of the functional (dual) one, which reads:
\begin{equation}\label{eq:mtd:min Ebij + Esmooth}
    \min\limits_{C, \Pi}\quad E_{\text{bij}}(\Pi, C) + \gamma\,E_{\text{smooth}}(\Pi)
\end{equation}
where $E_{\text{smooth}}$ penalizes non-smooth pointwise maps, its most basic version being the sum of the Dirichlet energies of the pointwise maps $E_{\text{smooth}}(\Pi) = \sum_{ij}E_D(\Pi_{ij})$ with $E_D$ being defined in Eq.\eqref{eq:Dirichlet pull back}. In section~\ref{sec:smoothness}, we highlight how other common energies for smoothness can be expressed as variations of this Dirichlet energy, thus enabling their straightforward introduction within our formulation.

\subsection{Smoothness-promoting Discrete Solver}\label{subsec:Solver}
We aim at solving Eq.~\eqref{eq:mtd:min Ebij + Esmooth} using a similar algorithm to the standard discrete solver discussed in Sec.~\ref{sec:background}. However as long as the energy $E_{\text{smooth}}$ includes quadratic terms in $\Pi_{ij}$, for instance the Dirichlet energy, this solver cannot be applied as it assumes row-separable variables (see Lemma 4.1 in~\cite{ren2021discrete}).
Since quadratic terms in the Dirichlet energy appear as $W$-norms of terms $\Pi_{ij}X_j$, we introduce auxiliary variables $Y_{ij}$ as surrogate for products $\Pi_{ij}X_{j}$, and add a corresponding coupling term between the two, resulting in a new \emph{coupled} smoothness energy:
\begin{equation}\label{eq:mtd:E_smooth coupled}
    E_{\text{sm}}^c(\Pi, Y) = E_{\text{smooth}}(\Pi, Y) + \beta\, \sum_{ij} \big\Vert Y_{ij} - \Pi_{ij} X_j \big\Vert^2_{A_i}
\end{equation}
where the second term is a spatial coupling term and, using some abuse of notations, $E_{\text{smooth}}(\Pi, Y)$ is obtained by replacing products $\Pi_{ij}X_j$ in $E_{\text{smooth}}(\Pi)$ by $Y_{ij}$.
In the particular case where $E_{\text{smooth}}=E_D$, the coupled smoothness energy is now \emph{row-separable} for $\Pi$:
\begin{equation}\label{eq:mtd:E_smooth couple Dirichlet}
    E_{\text{sm}}^c(\Pi, Y) = \sum_{ij} \big\Vert Y_{ij} \big\Vert^2_{W_i} + \beta\,  \big\Vert Y_{ij} - \Pi_{ij} X_j \big\Vert^2_{A_i}
\end{equation}

Note that this particular half-quadratic splitting was used in~\cite{RHM} to handle similar constraints. Furthermore we will show in Sec.~\ref{sec:smoothness} that multiple common energy for smoothness can benefit from this similar technique, resulting in a row-separable problem for $\Pi$  in all cases.

\begin{algorithm}[!b]
\DontPrintSemicolon
\SetKwData{Left}{left}\SetKwData{This}{this}\SetKwData{Up}{up}
\SetKwFunction{Union}{Union}\SetKwFunction{FindCompress}{FindCompress}
\SetKwInOut{Input}{Input}\SetKwInOut{Output}{Output}
\Input{Initial maps $\Scale[0.9]{\Pi_{12}^{\text{in}}, \Pi_{21}^{\text{in}}}$ and vertex positions $\Scale[0.9]{X_1, X_2}$}
\Output{Refined pointwise maps $\Scale[0.9]{\Pi_{12},\Pi_{21}}$}
\textbf{Initialization} : $\Scale[0.9]{\Pi_{ij}^{(0)} = \Pi_{ij}^{\text{in}}, \, Y_{ij}^{(0)} = \Pi_{ij}^{(0)}X_j}$ for $\Scale[0.9]{i,j\in\{1,2\}}$ 
 \While{Not converged}{
   $C^{(k+1)} = \argmin_C\; E_{\text{bij}}\big(\Pi^{(k)},C\big)$\;
   $Y^{(k+1)} = \argmin_Y\; E_{\text{sm}}^c\big(\Pi^{(k)},Y\big)$\;
   $\Pi^{(k+1)} =  \argmin_\Pi\; E_{\text{ours}}\big(\Pi,C^{(k+1)}, Y^{(k+1)}\big)$
  }    
\caption{Meta-algorithm}
\label{algo:Meta-algo}
\end{algorithm}

\paragraph*{Total energy} Eventually, the initial optimization problem, Eq.~\eqref{eq:mtd:min Ebij + Esmooth}, has been relaxed into a problem of the form $\min\limits_{\Pi, C, Y} E_{\text{ours}}(\Pi, C, Y) $ with
\begin{equation}
   E_{\text{ours}}(\Pi, C, Y) = E_{\text{bij}}\big( \Pi, C\big) + \gamma \, E_{\text{sm}}^c\big( \Pi, Y \big)
   \label{eq:energy_total}
\end{equation}
Crucially, this reformulation makes the total energy row-separable w.r.t. the pointwise maps $\Pi$.
%
%
We can therefore propose a general iterative method (summarized in Algorithm~\ref{algo:Meta-algo}) to minimize the total energy, in the spirit of the discrete solver, which iteratively updates each variable $\Pi, C, Y$ with the other two sets fixed.

\mypara{Solver} The solver described in Algorithm~\ref{algo:Meta-algo} is divided in three optimization problems for which we present the solution procedure.
\textbf{(1)} Computing $C^{(k+1)}$ from $\Pi^{(k)}$ reduces to a simple $K\times K$ linear system, which has actually been introduced as \emph{bijective ZoomOut} in~\cite{MapTree}.
\textbf{(2)} Computing $Y^{(k+1)}$ from $\Pi^{(k)}$ also reduces to a sparse linear system whose form depends on the choice of smoothness energy $E_{\text{smooth}}$, some of which are given in section~\ref{sec:smoothness}. In the case of $E_{\text{smooth}}=E_D$, computing $Y_{ij}$ requires solving $(W_i + \beta I_n)Y_{ij}=\beta\Pi_{ji}X_j$ where the system can be \emph{prefactored} to further improve efficiency.
\textbf{(3)} Since introducing auxiliary variables leads to a row-separable problem for $\Pi$, computing $\Pi^{(k+1)}$ from $C^{(k+1)}$ and $Y^{(k+1)}$ reduces to a simple nearest neighbor search. Note that this step is done in a high-dimensional space obtained by concatenating several terms, and can be heavily accelerated by only using coupling terms from equations Eq.~\eqref{eq:mtd:E_biject} and Eq.~\eqref{eq:mtd:E_smooth coupled}, which significantly reduces the embedding dimension on which to perform nearest neighbor
Finally, following~\cite{ren2021discrete}, we also increase the size $K$ of the functional map as iterations grow, which has shown to be a great regularization procedure in many spectral algorithms.

\section{Smoothness Analysis in Unified Framework}\label{sec:smoothness}


In this section, we formulate several existing formulations for promoting map smoothness, including non-rigid ICP (nICP)~\cite{nicp}, as-rigid-as-possible (ARAP)~\cite{arap}, reversible harmonic maps (RHM)~\cite{RHM}, and Smooth Shells~\cite{smoothshells}.
Our first objective is to provide a coherent formulation of various smoothness terms in the form of the Dirichlet energy on either a map or a deformation. Secondly, we aim to show how different energy terms and solvers can ultimately be introduced in our smoothness-promoting Discrete Solver. This will form the basis for our quantitative evaluation in the next section, in which we compare different terms within our solver.
We remain succinct regarding the following derivations and their incorporation in our algorithm, and refer the interested reader to the supplementary material for a more complete overview.

\mypara{nICP} was originally proposed to wrap a source shape $\Ss_1$ onto a target shape $\Ss_2$ via a per-vertex affine deformation field $\*D$.
nICP implicitly maintains a pointwise map $\Pi_{12}$ such that the deformed coordinates $\*D\circ X_1$ approximate the pointwise map $\Pi_{12}X_2$.
The total energy reads
\begin{equation}
    E_{\text{nicp}}(\Pi_{12}, \*D) = \big\Vert \*D \big\Vert^2_{W_1} + \beta\big\|\*D\circ X_1 - \Pi_{12}X_2\big\|^2_{A_1}
\end{equation}
with $\big\Vert \*D \big\Vert^2_{W_1} = \sum_{i \sim j} w_{ij}\big\Vert D_i - D_j \big\Vert_F^2$ extends the Dirichlet energy to per-vertex matrices.
In our algorithm, this energy may be used as a surrogate for $E_{\text{sm}}^c$ given in Eq.~\eqref{eq:mtd:E_smooth couple Dirichlet}.

\mypara{ARAP}is a commonly-used energy that aims at promoting \textit{local rigidity} of the shape deformation by enforcing the deformation to remain locally close to a rotation.
ARAP optimizes both for expected vertex coordinates $Y_{12}$ and per-vertex rotations $\*R$. The total reformulated energy reads:
\begin{equation}
    E_{\text{arap}}(Y_{12}, \*R) = \big\Vert Y_{12} \big\Vert^2_{W_1} + \lb  E_{\text{arap}}^{\text{rigid}}(Y_{12}, \*R),
\end{equation}
where $E_{\text{arap}}^{\text{rigid}}$ is a bilinear term promoting local rigid deformations. 
One can augment the energy using the coupling term from Eq.~\eqref{eq:mtd:E_smooth coupled} to use the ARAP energy in our algorithm.
\mypara{Smooth Shells} models the deformation $\*D$ as a simple per-vertex translation seen as a function $\Ss_1\to\RR^3$, restricted to lie in the \emph{spectral} basis of size $K$, i.e., $\*D\in\RR^{K\times 3}$. 
In addition smooth shells uses the ARAP energy to enforce the smoothness of the deformation.
Specifically if $Y_{12}=X_1 + \Phi_1\*D$ denotes the updated vertex positions, the shells energy is defined as
\begin{equation}
    E_{\text{shells}}(\*D, \*R, \Pi_{12}) = E_{\text{arap}}(Y_{12},\*R)
\end{equation}
which is augmented with a coupling term $\|X_1+\Phi_1 \*D - \Pi_{12}X_2\|^2_{A_1}$ to remain close to given correspondences.


\mypara{RHM} directly minimizes the Dirichlet energy of a map without manipulating deformation fields. To avoid making the map collapse the authors look for bijective maps with the lowest possible Dirichlet energy.
Specifically using notations of Sec.~\ref{sec:mtd}, smoothness is enforced by minimizing the same energy as in Eq.~\eqref{eq:mtd:E_smooth coupled} extended with a pointwise bijectivity term $\sum\limits_{ij} \|\Pi_{ij}Y_{ji} - X_i\|_{A_i}$, resulting in a slower solver.

\paragraph{} All these smoothness terms can be incorporated quickly within our solver, only affecting steps 2. and 3. of algorithm~\ref{algo:Meta-algo}. Furthermore note that, for fairness of comparison, we ignored additional building blocks used in these works like normal preservation, high-dimensional embeddings, etc. More details on these two points can be found in the supplementary material.

\section{Experiment}\label{sec:results}

    
    
    

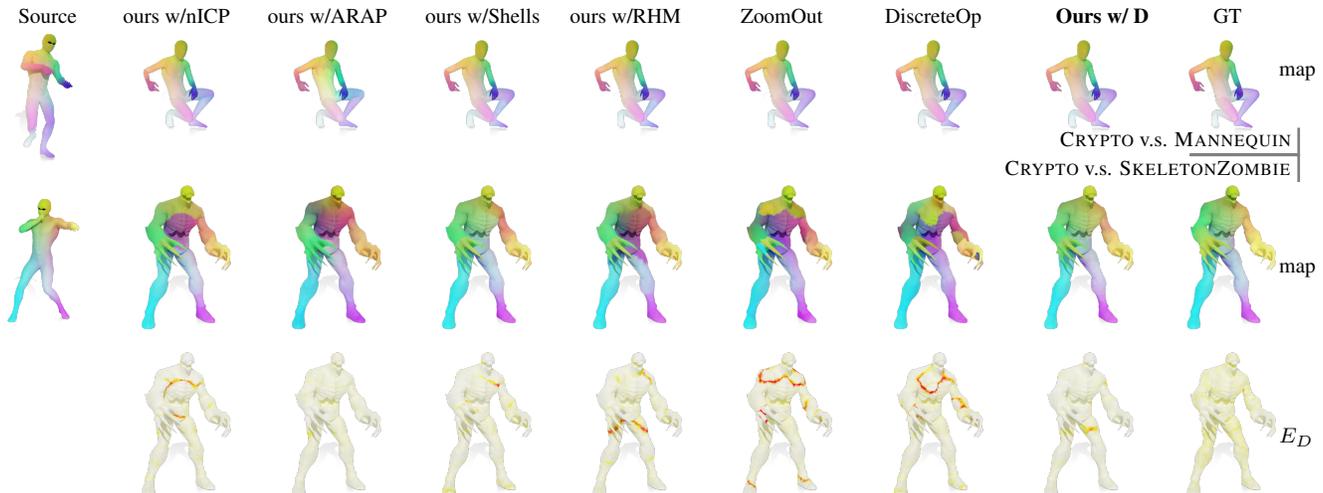
\begin{figure*}
    \centering
    \begin{overpic}[trim=0.9cm 0.3cm 0.35cm 0.8cm,clip,width=1\linewidth,grid=false]{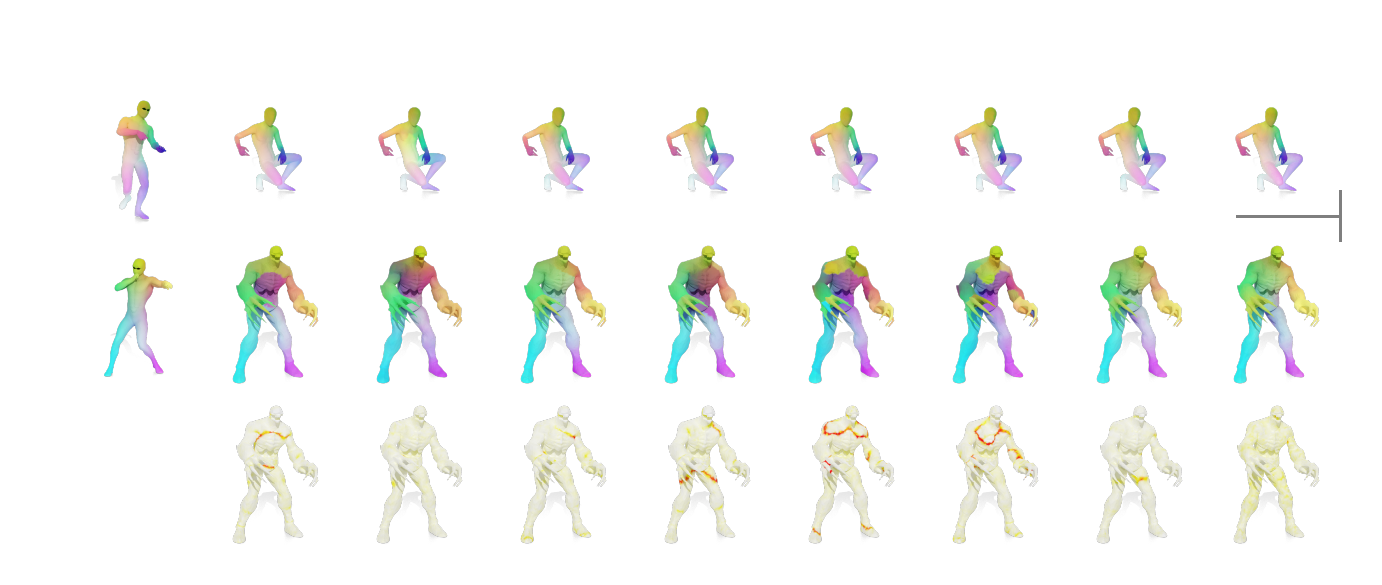}
    \put(81.3,27.6){\footnotesize \textsc{Crypto} v.s. \textsc{Mannequin}}
    \put(77.1,25.4){\footnotesize \textsc{Crypto} v.s. \textsc{SkeletonZombie}}
    
    \put(2,37){\footnotesize Source}
    \put(10,37){\footnotesize ours w/nICP}
    \put(21,37){\footnotesize ours w/ARAP}
    \put(33,37){\footnotesize ours w/Shells}
    \put(44,37){\footnotesize ours w/RHM}
    \put(57,37){\footnotesize ZoomOut}
    \put(68,37){\footnotesize DiscreteOp}
    \put(81,37){\footnotesize\bfseries Ours w/ D}
    \put(93,37){\footnotesize GT}
    
    \put(98,33){\footnotesize map}
    \put(98,18){\footnotesize map}
    \put(98,5){\footnotesize $E_D$}
    
    \end{overpic}\vspace{-9pt}
    \caption{Qualitative evaluation on two pairs from \textsc{DeformThings4D-Matching}. For a near-isometric shape pair shown on the \emph{top}, all methods achieve smooth maps. For a shape pair that is far from isometry shown on \emph{bottom}, nICP, ARAP, RHM, and Shells achieve relatively smooth maps but contain large patch of back-to-front ambiguity. The maps obtained by ZoomOut and Discrete Solver are locally smooth due to their spectral representation, but fail to maintain global smoothness. As a comparison, our methods can be robustly generalized to non-isometric shape maps and achieve globally smooth maps.}
    \label{fig:res:deform}
\end{figure*}

\subsection{\textsc{DeformThings4D-Matching} Dataset}
We propose \textsc{DeformThings4D-Matching} (Fig.~\ref{fig:res:dt_deform4d}),  a new dataset based on the \textsc{DeformThings4D} dataset~\cite{li20214dcomplete}, a rich synthetic dataset with significant variations in both identities and types of motions, containing 1,972 animation sequences spanning 31 categories of humanoids and animals.
However, using \textsc{DeformThings4D} to evaluate shape matching is difficult since: (1) most shape models are disconnected; (2) the meshes belong to the same category are in the same triangulation, which provides perfect ground-truth but can lead to over-fitting issues for matching algorithms~\cite{ren2018continuous}, while cross-category ground truth is missing; (3) some meshes of the synthesized poses have unrealistic distortions such as large self-intersections and unnatural twists.
We therefore select 56 animal categories and 8 humanoid categories from \textsc{DeformThings4D}, each containing 15-50 poses selected from different motion clips while ensuring large enough pose variations. 
We then apply LRVD~\cite{yan2014low} to \emph{independently} remesh all the meshes in the same category. 
For the humanoid shapes, we further use the commercial software \footnote{https://www.russian3dscanner.com/}{\textsc{R3DS}} to non-rigidly fit one shape into another to get \textit{cross-category} correspondences. See Fig.~\ref{fig:res:dt_deform4d} for some examples, where the corresponding vertices are assigned the same color.
See supplementary materials for more details of how we construct the dataset and obtain the ground-truth correspondences between the remeshed shapes with different triangulations. The dataset is available at \url{https://github.com/llorz/3DV22_DeformingThings4DMatching_dataset}.

\subsection{Comparison on Smoothness Formulation}
We evaluate our method on the standard benchmark for non-isometric shape matching TOSCA non-Isometric Dataset~\cite{TOSCA}, and the cross-category humanoid shape pairs from our \textsc{DeformThings4D-Matching} Dataset.
Note that on standard benchmarks like the FAUST dataset~\cite{FAUST}, existing methods already perform well as shapes remain near-isometric. We provide some results in Table~\ref{tab:res:faust} to show our method performs similarly in these simple cases, and refer to supplementary material for additional discussions.

\mypara{Evaluation Metrics}
We follow~\cite{ren2018continuous} to measure the \emph{accuracy}, \emph{bijectivity}, \emph{coverage} and \emph{runtime} to compare different methods.
Additionally, We apply Eq.~\eqref{eq:Dirichlet pull back} to compute the Dirichlet energy on the obtained pointwise maps to evaluate the \emph{smoothness}. See supplementary materials for detailed definitions and discussions.

\begin{table}[!t]
\centering
\caption{\textsc{DeformThings4D-Matching} Dataset: Summary over 433 shape pairs. We highlight the best two in blue, except those of Shells and RHM (see text for details).} \label{tab:res:deform summary}
\footnotesize
{\def\arraystretch{0.95}\tabcolsep=0.3em
\begin{tabular}{cccccc}\toprule[0.8pt]
methods & \bfseries\itshape accuracy & \bfseries\itshape bijectivity & \bfseries\itshape smoothness & \bfseries\itshape coverage  &\bfseries\itshape runtime (s)\\ \midrule[0.8pt]
Init & $12.71$  & $11.70$  & $3.60$ & $24.57$\% & - \\ \midrule[0.5pt]
RHM & $11.8$ & $1.6$ &  $0.50$ & $56.6$\%  \\
Shells & $11.4$ & $5.1$   & $1.50$ & $50.8$\%\\\midrule[0.5pt]
Ours w/ ARAP & $12.16$  & $11.70$  & \cellcolor{tabblue!20} $0.71$ & $31.0$\% & $25.3$ \\
Ours w/ nICP & $9.56$  & $3.89$  & $1.72$ & $40.4$\% & $100.8$\\
Ours w/ Shells & $8.41$  & $2.59$  & $2.18$ & $51.7$\% & $48.2$ \\ \midrule[0.5pt]
ZO & $8.57$  & $7.14$  & $4.02$ & \cellcolor{tabblue!20} $67.0$\% &  \cellcolor{tabblue!20}  $17.5$\\
DO & $9.01$  &\cellcolor{tabblue!20} $1.78$  & $3.21$ & \cellcolor{tabblue!20} $62.4$\% & $40.9$\\ \midrule[0.5pt]
Ours w/ D & \cellcolor{tabblue!20}$8.19$  & $2.63$  & $1.56$ & $50.4$\% & \cellcolor{tabblue!20}21.4 \\
Ours w/ RHM & \cellcolor{tabblue!20} $8.10$  &\cellcolor{tabblue!20} $2.18$  & \cellcolor{tabblue!20}$1.47$ & $56.0$\% & $42.1$\\ \bottomrule[0.8pt]

\end{tabular}}
\end{table}

\mypara{Initialization \& Baselines}
Since the tested shape pairs are highly non-isometric and challenging, standard shape descriptors failed to produce reasonable initialization as shown in supplementary.
We therefore compute each initial map from a $5\times 5$ functional map obtained by using $5$ landmarks.
%
Our baselines can be categorized into three groups: 
(1) We compare to ZoomOut (ZO)~\cite{zoomout} and Discrete Solver (DO)~\cite{ren2021discrete}, the current-state-of-the-art refinement methods in functional maps pipeline. 
(2) We compare the standard Dirichlet Energy with the different variants presented in Sec.~\ref{sec:smoothness}, namely nICP~\cite{nicp}, ARAP~\cite{arap}, Shells~\cite{smoothshells} and RHM~\cite{RHM}, all using the same algorithm. We highlight the Dirichlet energy (ours w/ D) and the RHM energy (ours w/ RHM) as respectively the simplest and globally best performing energies within our algorithm, which we both advocate.
(3) We also include the results using original implementations of RHM and Shells for reference only since additional regularizers besides smoothness are included.


\begin{table}[!t]
\centering
\caption{Results on a random subset of 200 pairs of the FAUST dataset. We highlight the best two in blue.} \label{tab:res:faust}
\footnotesize
{\def\arraystretch{0.95}\tabcolsep=0.85em
\begin{tabular}{ccccc}\toprule[0.8pt]
methods & \bfseries\itshape accuracy & \bfseries\itshape bijectivity & \bfseries\itshape smoothness & \bfseries\itshape coverage \\ \midrule[0.8pt]
Init & $6.45$  & $5.51$  & $2.67$ & $38.47$ \%\\ \midrule[0.5pt]
ZO & \cellcolor{tabblue!20}$3.95$  & $2.16$  & $0.79$ & \cellcolor{tabblue!20}$82.16$ \% \\
DO  & $4.07$  & \cellcolor{tabblue!20}$1.08$  & $0.86$ & $77.96$ \% \\ \midrule[0.5pt]
Ours w/ D & $4.43$  & $1.83$  & \cellcolor{tabblue!20}$0.64$ & $67.47$ \% \\
Ours w/ RHM & \cellcolor{tabblue!20}$3.94$  & \cellcolor{tabblue!20}$1.11$  & \cellcolor{tabblue!20}$0.71$ & \cellcolor{tabblue!20}$79.26$ \% \\
\bottomrule[0.8pt]
\end{tabular}}
\end{table}

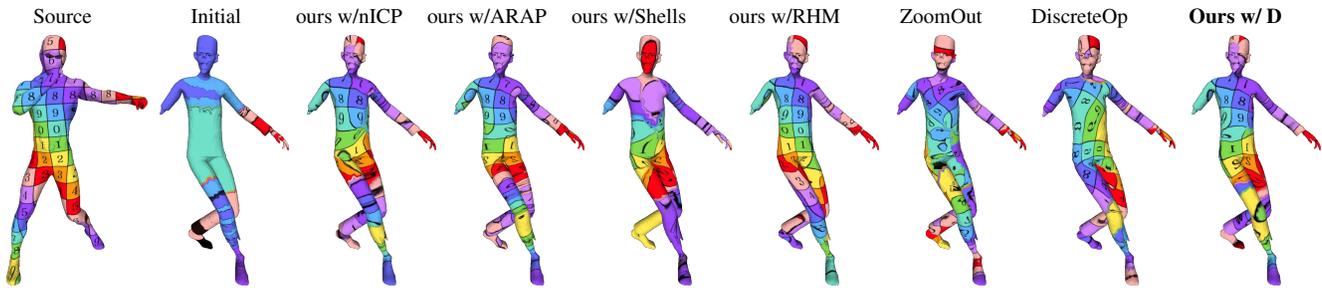
\begin{figure*}[!t]
    \centering
    \begin{overpic}[trim=0cm 0cm 0cm 0cm,clip,width=1\linewidth,grid=false]{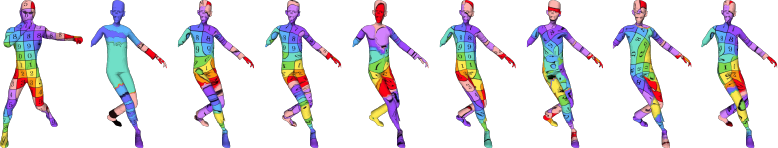}
    \put(2,20){\footnotesize Source}
    \put(14,20){\footnotesize Initial}
    \put(22,20){\footnotesize ours w/nICP}
    \put(32,20){\footnotesize ours w/ARAP}
    \put(43,20){\footnotesize ours w/Shells}
    \put(55,20){\footnotesize ours w/RHM}
    \put(68,20){\footnotesize ZoomOut}
    \put(78,20){\footnotesize DiscreteOp}
    \put(90,20){\footnotesize\bfseries Ours w/ D}
    \end{overpic}
    \caption{Starting from a poor initial map, our method can produce a more smooth and accurate map compared to the baseline methods.}
    \label{fig:exp:texture}
\end{figure*}

\mypara{DeformThings4D-Matching Dataset}
We report the average metrics over 433 cross-category shape pairs from the humanoid shapes from our \textsc{DeformThings4D-Matching} dataset in Tab.~\ref{tab:res:deform summary}.
Among all the baseline methods, our method achieves the best accuracy.
Compared to ZoomOut (ZO) and Discrete Solver (DO), our two selected energies achieve $3\times$ better smoothness on average with comparable bijectivity and coverage. It suggests that, our method, as an extended algorithm of discrete solver by adding a smoothness term, is indeed effective to promote map smoothness. In supplementary, we also report \emph{per-category} map evaluation.
We show two qualitative examples in Fig.~\ref{fig:res:deform}, where the obtained maps are visualized by color transfer.
For the pair between \textsc{Crypto} and \textsc{SkeletonZombie}, we also visualize the per-vertex smoothness error for each map. We additionally display texture transfer for a difficult pair in Fig.~\ref{fig:exp:texture}, using~\cite{ezuz2017deblurring} to obtain a vertex-to-point map for each method to improve visualization. While this figure shows that our maps clearly outperform standard spectral method starting from poor initialization, there is room for improvement for all energies.
%


\begin{table}
\centering
\caption{TOSCA Non-Isometric Dataset: Summary over 95 shape pairs. We highlight the best two in blue, except those of Shells and RHM (see text for details).} \label{tab:res:tosca}
\footnotesize
{\def\arraystretch{1}\tabcolsep=0.3em
\begin{tabular}{cccccc}\toprule[0.8pt]
methods & \bfseries\itshape accuracy & \bfseries\itshape bijectivity & \bfseries\itshape smoothness & \bfseries\itshape coverage  &\bfseries\itshape runtime (s)\\ \midrule[0.8pt]
Init & $7.51$  & $7.23$  & $1.94$ & $26.9$\% & - \\ \midrule[0.5pt]
RHM & $9.20$  & $1.37$  & $1.55$ & $54.3$ \% & $818$\\
Shells & $10.20$  & $6.72$  & $5.58$ & $45.6$ \% &  $29.0$\\ \midrule[0.5pt]
ours w/ ARAP & $7.55$  & $8.35$  &\cellcolor{tabblue!20} $0.83$ & $48.6$\% & $42.8$\\
Ours w/ nICP & $7.78$  & $3.63$  & \cellcolor{tabblue!20} $1.16$ & $40.2$\% & $178$\\
Ours w/ Shells & $11.85$  & $7.40$  & $1.18$ & $37.8$\%  & $72.5$\\ \midrule[0.5pt]
ZO & $12.47$  & $8.17$  & $6.53$ & \cellcolor{tabblue!20} $56.8$\% & \cellcolor{tabblue!20} $33.7$ \\
DO & $13.30$  & \cellcolor{tabblue!20} $1.90$  & $5.51$ & \cellcolor{tabblue!20} $53.4$\% & $79.2$\\ \midrule[0.8pt]
Ours w/ D & \cellcolor{tabblue!20} $7.25$  & $3.02$  & $1.22$ & $42.2$\% & \cellcolor{tabblue!20} $33.3$ \\
Ours w/ RHM& \cellcolor{tabblue!20} $6.26$  &\cellcolor{tabblue!20} $1.87$  & $1.39$ & $53.1$\% & $40.1$\\ \bottomrule[0.8pt]
\end{tabular}}
\end{table}
\mypara{TOSCA Non-Isometric Dataset} contains cross-category correspondences among one gorilla shape (with 5 different poses), one male shape (with 7 different poses), and one female shape (with 12 different poses). 
We use all 95 non-isometric shape pairs between the gorilla shapes and the human (male and female) shapes. The summary evaluation is shown in Tab.~\ref{tab:res:tosca}. 
See supplementary for qualitative examples.
Enforcing the smoothness of the pointwise maps via Dirichlet energy (Ours + D) help us achieve much more accurate and $5\times$ smoother maps.
We additionally highlight that adding extra pointwise bijectivity (ours w/ RHM) has a positive effect on the metrics, but results in a slower solver. Finally, while ARAP and nICP energies perform quite well, the Shells energy seems to suffer from the high level of non-isometry in the dataset as it mainly relies on spectral quantities.


\subsection{Implementation \& Parameters}
We implemented all the baselines (based on their released code) and our methods in Python to guarantee a fair comparison. 
We follow the discrete solver~\cite{ren2021discrete} to adopt the progressive upsampling technique into our algorithm, which is introduced in~\cite{zoomout}, and gradually increase the spatial coupling term weight $\gamma$ to avoid over-smoothing in the earlier iterations. Detailed parameters can be found in supplementary, or in the released version of the code at \url{https://github.com/RobinMagnet/smoothFM}.
%
%

\section{Conclusion, Limitations \& Future Work}\label{sec:conclusion}
In this work, we extended the discrete solver formulation from \cite{ren2021discrete} to optimize the Dirichlet energy to promote map smoothness. We then proposed an efficient algorithm that can produce high-quality and smooth maps from noisy initial maps for between non-isometric surfaces.
%
Furthermore, we demonstrated that multiple previously proposed methods for computing smooth maps, including nICP \cite{nicp}, ARAP \cite{arap}, RHM \cite{RHM}, and Smooth Shells \cite{smoothshells}, can be reformulated in a coherent framework.
This allowed us to compare and analyze different formulations for smoothness using a single algorithm.
Finally, to address the scarcity of evaluation data, we proposed a new dataset based on \textsc{DeformThings4D}, with variable mesh structure, and dense ground truth cross-category correspondences for eight challenging categories. We believe both our framework and this dataset can be helpful for the shape matching community.

Our method still has some limitations. First, optimizing the Dirichlet energy can indeed greatly improve the smoothness compared spectral methods. This, however, can come at the expense of loss of coverage, and we observe that our maps can still collapse locally, as seen from the texture transfer of Figure~\ref{fig:exp:texture}. It would be interesting to investigate techniques that to further prevent \textit{local collapse} and obtain a smooth maps with high coverage.
Second, our results show that the proposed method improves significantly results from ZoomOut and discrete solver on complete shapes, even for non-isometric cases. However, for the partial matching setting, though our maps still outperform ZoomOut and discrete solver, there is still a lot room for further improvement. 
Finally, our energy is a weighted sum of a bijectivity and a smoothness term, which can become hard to balance across all initialization quality.


%
In the future, we would like to study different energies for partial matching  and to ways prevent local map collapse. 
It will also be interesting to apply our approach for computing dense correspondences in other domains, such as point clouds, graphs or even 2D images.

\mypara{Acknowledgments}
The authors thank the anonymous reviewers for their valuable comments and suggestions. Parts
of this work were supported by the ERC Starting Grant No.
758800 (EXPROTEA), the ERC Consolidator Grant No. 101003104 (MYCLOTH), and the ANR AI Chair AIGRETTE.

{\small
\bibliographystyle{ieee_fullname}
\bibliography{3DV_a_main}
}



\pagebreak

\setcounter{equation}{0}
\setcounter{figure}{0}
\setcounter{table}{0}
\setcounter{page}{1}
\setcounter{section}{0}
\makeatletter
\renewcommand{\theequation}{S\arabic{equation}}
\renewcommand{\thefigure}{S\arabic{figure}}
\newcommand{\bibnumfmt}[1]{[S#1]}
\newcommand{\citenumfont}[1]{S#1}
\twocolumn[{{\Large \bf  
 Supplemental Materials: Smooth Non-Rigid Shape Matching via Effective Dirichlet Energy Optimization\par}}
 \medbreak
 \par\vspace{0.3cm}]

\section{Smoothness Reformulation}\label{subsec:smoothness reformulation}
In this section, we give details on the reformulation of smoothness methods provided in the main manuscript.

\subsection{Non-Rigid ICP}
Non-Rigid ICP (nICP)~\cite{nicp} deforms a source shape $\Ss_1$ into a target shape $\Ss_2$ using a per-vertex affine deformation $\*D$. Global energy reads, trying to fit a pointwise map $\Pi_{12}$
\begin{equation}
    E_{\text{nicp}}(\Pi_{12}, \*D) = \big\Vert \*D \big\Vert^2_{W_1} + \beta\big\|\*D\circ X_1 - \Pi_{12}X_2\big\|^2_{A_1}
\end{equation}
with $\big\Vert \*D \big\Vert^2_{W_1} = \sum_{i \sim j} w_{ij}\big\Vert D_i - D_j \big\Vert_F^2$ and $\*D\circ X_1$ the deformed vertex coordinates.
Given a point-wise map $\Pi_{12}$, one can directly incorporate this energy in our algorithm, where solving for $Y_{12}$ is replaced by solving for $\*D$, and then setting $Y_{12}=\*D\circ  X_1$. Solving for $\*D$ reduces to a simple linear system, as explained in~\cite{nicp}.
Note that in the original work, nICP algorithm uses graph Laplacian instead of cotan Laplacian, but we find that using cotan weights is more stable in the case of triangle meshes. We furthermore ignored landmarks preservation terms, borders skipping heuristic, normals preservation and self-intersection verification procedures for simplicity.


\subsection{As-Rigid-As-Possible}
As-Rigid-As-Possible (ARAP)~\cite{arap} promotes \textit{local rigidity} of the deformation of a shape $\Ss_1$ using per-vertex rotations $\*R$, which results in minimizing the following energy:
\begin{equation}
    E_{\text{arap}}(\*R, Y) = \sum_{i\sim j} w_{ij} \big\Vert (y_i - y_j) - R_i (x_i - x_j) \big\Vert_F^2
\end{equation}
where $y_i$ are the expected vertex coordinates and $x_i$ the undeformed coordinates.

%
We observe that the ARAP energy can be decomposed into two main components including a smoothness term and a rigidity term: 
\begin{align}
    E_{\text{arap}}(\*R, Y) & = E_{\text{arap}}^{\text{smooth}}(Y) - 2E_{\text{arap}}^{\text{rigid}}(\*R, Y) + \text{const.}\\
    E_{\text{arap}}^{\text{smooth}}(Y) & = \sum_{(x_i, \,x_j)\in\Ee(\Ss_1)} w_{ij} \big\Vert y_i - y_j\big\Vert_F^2\\
    E_{\text{arap}}^{\text{rigid}}(\*R, Y) & = \sum_{(x_i, \, x_j)\in\Ee(\Ss_1)} w_{ij} (y_i - y_j)^T R_i (x_i - x_j)
\end{align}
with $E_{\text{arap}}^{\text{smooth}} = \big\Vert Y \big\Vert^2_{W_1} = E_D(Y)$. Note, however, that the default ARAP energy does not have a coupling term to ensure that $Y$ remains on the surface of $\Ss_2$. Therefore, to avoid a trivial solution, such an energy must rely on pre-existing landmarks to make sure that the deformation maps onto the target shape.
In our algorithm, given a pointwise map $\Pi_{12}$, we instead decide to add a coupling term between the expected coordinates $Y_{12}$ and transferred coordinates $\Pi_{12}X_1$, which slightly modifies the linear system to solve when minimizing over $Y_{12}$, but doesn't involve the rotations $\*R$. Therefore, given a pointwise map $\Pi_{12}$, one first needs to compute local rotations $\*R$ and can then obtain the expected coordinates $Y_{12}$ by solving a linear system.

\subsubsection{Smooth Shells}
Smooth Shells~\cite{smoothshells} models the deformation $\*D$ as a simple per-vertex translation seen as a function $\Ss_1\to\RR^3$, which is restricted to lie in the \emph{spectral} basis of size $K$, i.e., $\*D\in\RR^{K\times 3}$. 
In addition smooth shells uses the ARAP energy to enforce the smoothness of the deformation, which therefore adds additional local rotation $\*R$.
Specifically, $X_1 + \Phi_1\*D$ would give the updated vertex positions and the smoothness is then defined as:
\begin{equation}
   E_{\text{shells}}^{\text{smooth}}(\*D, \*R) =  E_{\text{arap}}(\*R, X_1 + \Phi_1\*D)
\end{equation}
The smoothness energy is again associated with a coupling term which ensures the deformed shape remains close to the current correspondences
$\|X_1+\Phi_1 \*D - \Pi_{12}X_2\|^2_{A_1}$. Note that in the original work, vertices $X_1$ and $X_2$ are also projected to a spectral basis, and extra feature and normal preservation terms are added. In practice, solving for $\*D$ reduces to solving a $K\times K$ linear system, compared to the $n\times n$ linar system obtained with standard ARAP.

\subsubsection{Reversible Harmonic Maps}
Reversible Harmonic Maps (RHM)~\cite{RHM} directly minimizes the Dirichlet energy of a map without manipulating deformation fields. To avoid making the map collapse the authors look for bijective maps with the lowest possible Dirichlet energy. Vertices of the pull-back shape $\Pi_{ij}X_j$ for $(i,j)\in\left\{(1,2), (2,1) \right\}$ are again estimated via an auxiliary variable $Y_{ij}$ and the energy reads as the sum in both directions of $E_{\text{rhm}}^{\text{half}}$ with :
\begin{equation}
\begin{split}
    E_{\text{rhm}}^{\text{half}}\left(\Pi_{ij},\Pi_{ji},Y_{ij},Y_{ji}\right) =&\ E_D(Y_{ij}) + E_{\text{rhm}}^{\text{bij}}(\Pi_{ji}, Y_{ij})\\ &+ E_{\text{rhm}}^{\text{couple}}(\Pi_{ij}, Y_{ij}).
\end{split}
\end{equation}
Here, again, we recognize the Dirichlet energy of the estimated map $E_D(Y_{ij})$, and two terms $E_{\text{rhm}}^{\text{bij}}$ and $E_{\text{rhm}}^{\text{couple}}$ which respectively enforce bijectivity and coupling: 
\begin{align}
    E_{\text{rhm}}^{\text{bij}}(\Pi_{ji}, Y_{ij}) &= \big\Vert \Pi_{ji}Y_{ij} - X_j\big\Vert^2_{A_j} \\
    E_{\text{rhm}}^{\text{couple}}(\Pi_{ij}, Y_{ij}) &= \big\Vert Y_{ij} - \Pi_{ij}X_j\big\Vert_{A_i}
\end{align}
This formulation leads to a computationally expensive iterative solver, which can obtain to great results given an already good initialization. Additionally, the authors use a high-dimensional embedding obtained via MDS~\cite{cox2008multidimensional} which mimics the geodesic distance, instead of directly using the embedding coordinates.

\section{\textsc{DeformThings4D-Matching} Dataset}\label{sec:appendix:dataset}
\begin{figure}[!t]
    \centering
    \begin{overpic}[trim=0.8cm 0.4cm 0cm 0.2cm,clip,width=1\linewidth,grid=false]{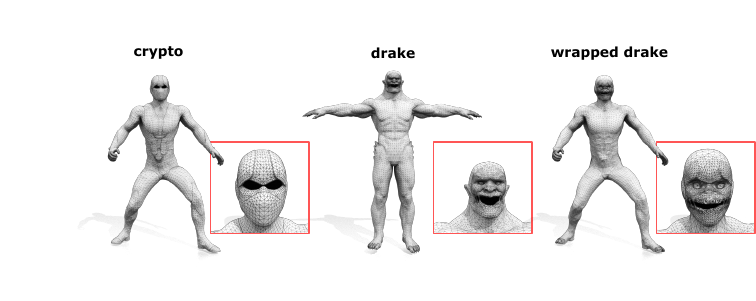}
    \end{overpic}\vspace{-9pt}
    \caption{Example of wrapping a \textsc{Drake} shape to a \textsc{Cryto} shape to establish cross-category correspondences.}
    \label{fig:appendix:eg_drake}
\end{figure}

Here we discuss in details how we construct our dataset from the \textsc{DeformThings4D}~\cite{li20214dcomplete} for shape matching task:

\begin{enumerate}[leftmargin=*]
    \item \textbf{Select Models.} We first pick models in \textsc{DeformThings4D} that are close to watertight. Specifically, we only keep the models where the number of vertices in the largest connected components is more than 75\% of the total number of vertices. Then the largest connected component is taken if the model is disconnected. As a result, we get 56 animal models and 8 humanoid models.
    \item \textbf{Select Poses.} For each watertight model, we collect all motion clips in \textsc{DeformThings4D} and select poses from all the frames that are sufficiently different from each other. Specifically, we first pick a base pose that is close to an A-pose: we find the pose that has large range in $z$-axis and has relatively small range in $xy$-axis. We then recursively find new pose from the collection that have the largest difference in vertex positions to the chosen ones, until we get 50 poses or all the poses are included. We then manually check each chose pose and remove unrealistic poses with large distortion or self-intersection. As a result, the number of poses for each model has a range from 30 to 50.
    \item \textbf{Remeshing.} The chosen poses in each model are in the same triangulation, which can lead to overfitting issues for some shape matching methods~\cite{ren2018continuous}. We therefore apply a geometry-aware remeshing algorithm, LRVD~\cite{yan2014low}, to independently remesh all the poses to the resolution of around 8K vertices. The correspondences between the remeshed shapes are propagated by nearest-neighbor searching between the remeshed shapes and the original shapes. To fix the potential topological errors in the nearest neighbor map, we apply spectral ICP~\cite{ovsjanikov2012functional} at dimension 500 of the Laplace-Beltrami Basis.
    \item \textbf{Wrapping.} We also provide cross-category correspondences for the 8 humanoid models. Specifically, we use the commercial software R3DS to wrap the rest 7 models (\textsc{Zlorp, Mannequin, Drake, Ninja, Prisoner, PumpkinHulk, SkeletonZombie}) to the chosen model (\textsc{Crypto}, the left-most shapes in Fig.~2 in the main paper). For each pair, we manually select 50-80 landmarks on shapes for wrapping. Note here we wrap the original models and propagate the correspondences to the remeshed shapes afterwards. Specifically, the cross-category correspondences among the original poses can be established by nearest-neighbor searching between the wrapped shape and the target shape (see Fig.~\ref{fig:appendix:eg_drake} for an example of a wrapped shape), which are then propagated to the remeshed poses similar to step 3. Note that, since some shapes are far from isometry or even incomplete, the wrapped results are not perfect, and hence the established correspondences via map compositions can be inaccurate. In general, as illustrated in Fig.~2 in the main paper, the established correspondences are in reasonable accuracy.  
\end{enumerate}

\section{FAUST dataset}
\label{sec:appendix:Faust}
The FAUST dataset~\cite{FAUST} consists in 100 meshes of 10 individuals in 10 different poses.

This dataset is used as a standard benchmark for most shape-matching algorithms. However as all shapes are near-isometric, many methods achieve smooth and accurate results for this dataset. This therefore gives very little room for improvement regarding the smoothness.

We provide results on a random subset of 200 pairs in the main manuscript, where pairs we selected so that only cross-individual ones are considered.

\section{Additional Results}\label{sec:appendix:res}

\begin{table}[!t]
\centering
\caption{\textbf{Accuracy} on \textsc{DeformThings4D-Matching}}
\label{tab:append:deform_acc}
\footnotesize
{\def\arraystretch{1.05}\tabcolsep=0.15em
\begin{tabular}{ccccccc}\toprule[0.8pt]
\multirow{2}{*}{methods} & \multicolumn{4}{c}{\textbf{\textit{near-isometric}}} & \textbf{\textit{partial}} & \textbf{\textit{non-iso}} \\ \cmidrule(lr){2-5}\cmidrule(lr){6-6}\cmidrule(lr){7-7}
 & \textsc{Zlorp} & \textsc{Drake} & \textsc{Mannequin} & \textsc{Ninja} & \textsc{Prisoner} & \textsc{Zombie}  \\ \midrule[0.8pt]
 Init & $11.49$ & $9.59$  & $8.62$  & $10.43$  & $20.78$  & $15.33$   \\ \midrule[0.5pt]
Ours w/ ARAP & $11.22$ & $9.04$  & $8.10$  & $9.88$  & $19.91$  & $14.83$    \\
Ours w/ nICP  & $7.29$ & $7.07$  & $4.61$  & $5.25$  & $21.18$  & $11.95$  \\
Ours w/ Shells & $3.25$ & $7.78$  & $4.11$  & $4.73$  & $20.27$  & $10.32$    \\ \midrule[0.5pt]
ZO & $3.43$ & $5.74$  & $3.33$  & $4.61$  & $20.59$  & $13.71$    \\
DO & $3.26$ & $5.95$  & $3.64$  & $5.10$  & $19.59$  & $16.53$    \\ \midrule[0.5pt]
Ours w/ D & $3.72$ & $6.93$  & $4.18$  & $4.80$  & $19.81$  & $9.71$    \\
Ours w/ RHM & $3.70$ & $5.63$  & $3.94$  & $5.46$  & $18.85$  & $11.00$    \\\bottomrule[0.8pt]
\end{tabular}}
\end{table}

\begin{table}[!t]
\caption{\textbf{Bijectivity} on \textsc{DeformThings4D-Matching}}\label{tab:append:deform_biject}
\centering
\footnotesize
{\def\arraystretch{1.05}\tabcolsep=0.15em
\begin{tabular}{ccccccc}\toprule[0.8pt]
\multirow{2}{*}{methods} & \multicolumn{4}{c}{\textbf{\textit{near-isometric}}} & \textbf{\textit{partial}} & \textbf{\textit{non-iso}} \\ \cmidrule(lr){2-5}\cmidrule(lr){6-6}\cmidrule(lr){7-7}
 & \textsc{Zlorp} & \textsc{Drake} & \textsc{Mannequin} & \textsc{Ninja} & \textsc{Prisoner} & \textsc{Zombie}  \\ \midrule[0.8pt]
 Init & $11.69$ & $7.17$  & $6.58$  & $10.69$  & $22.53$  & $11.52$  \\\midrule[0.5pt]
Ours w/ ARAP & $11.93$ & $7.25$  & $7.69$  & $10.42$  & $21.71$  & $11.18$    \\
Ours w/ nICP & $3.63$ & $2.73$  & $2.58$  & $2.49$  & $7.17$  & $4.71$    \\
Ours w/ Shells & $1.67$ & $2.16$  & $2.22$  & $2.23$  & $3.56$  & $3.71$    \\\midrule[0.5pt]
ZO & $2.14$ & $4.05$  & $1.37$  & $3.99$  & $21.19$  & $10.11$    \\
DO & $1.27$ & $1.55$  & $1.63$  & $1.46$  & $2.26$  & $2.52$   \\ \midrule[0.5pt]
Ours w/ D & $1.77$  & $2.12$  & $2.30$  & $2.25$  & $3.60$  & $3.74$   \\
Ours w/ RHM & $1.42$ & $1.84$  & $1.82$  & $1.94$  & $2.81$  & $3.24$    \\\bottomrule[0.8pt]
\end{tabular}
}
\end{table}

We evaluate different methods using accuracy, bijectivity, coverage, and smoothness of the maps as metrics. We also report runtime to compare the efficiency. 
Specifically, we compute the geodesic distances between the obtained maps $T_{ij}$ and the ground-truth maps (if available) to measure the \emph{accuracy} (see Tab.~\ref{tab:append:deform_acc}). 
Similarly, we compute the geodesic distances between the composite maps $T_{ij}\circ T_{ji}$ and the identity map $I_{n_i}$ to measure the \emph{bijectivity} of the pointwise maps (see Tab.~\ref{tab:append:deform_biject}). 
We compute the Dirichlet energy on the obtained pointwise maps to evaluate the smoothness (defined in Eq.~(3) in the main paper) as shown in Tab.~1 in the main paper. Here we additionally evaluate the conformal distortion~\cite{RHM,MapTree}, another popular smoothness metric as shown in Tab.~\ref{tab:append:deform_smooth}.
We finally compute coverage of a pointwise map $T$, i.e., the area ratio of the target shape that is covered by map $T$, which evaluates the map surjectivity (see Tab.~\ref{tab:append:deform_cov}). This metric must be considered in pair with smoothness to detect degenerate case of trivial maps with perfect smoothness.
For example, a trivial map where are vertices on the source are mapped to the same vertex on the target, is perfectly smooth w.r.t. the Dirichlet energy, but its coverage is close to zero.
Therefore, in the ideal case, the best map is the one with zero Dirichlet energy and 100\% coverage.
All metrics are reported as an average over all the tested shape pairs.

\begin{figure*}
    \centering
    \begin{overpic}[trim=0.7cm 0.5cm 1.2cm 0.6cm,clip,width=1\linewidth,grid=false]{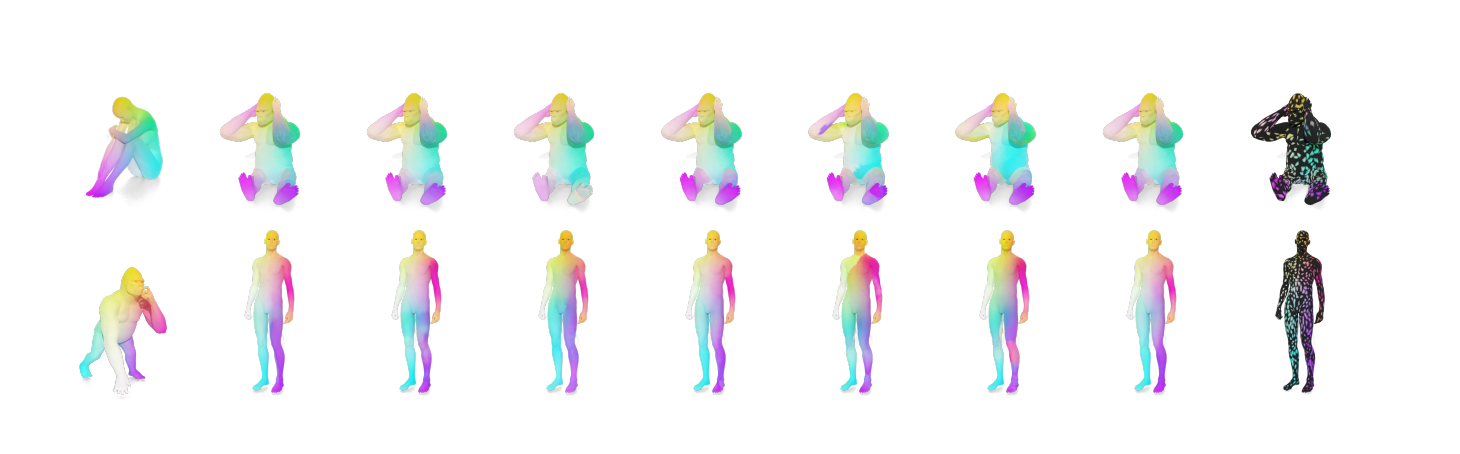}
    \put(2,25){\footnotesize Source}
    \put(12,25){\footnotesize ours w/nICP}
    \put(22,25){\footnotesize ours w/ARAP}
    \put(34,25){\footnotesize ours w/Shells}
    \put(45.5,25){\footnotesize ours w/RHM}
    \put(58.5,25){\footnotesize ZoomOut}
    \put(69.5,25){\footnotesize DiscreteOp}
    \put(81.5,25){\footnotesize\bfseries Ours w/ D}
    \put(95.5,25){\footnotesize GT}
    \end{overpic}\vspace{-3pt}
    \caption{We show two non-isometric shape pairs from TOSCA dataset can compare pointwise maps obtained from different methods via color transfer. Note that TOSCA non-isometric dataset only provide \emph{sparse} ground-truth correspondences. We therefore color the vertices that do not have GT correspondences in \emph{black}.}
    \label{fig:res:tosca}

\end{figure*}

In Fig.~\ref{fig:res:tosca}, we show some qualitative results on the TOSCA non-isometric dataset.


\begin{table}[!t]
\centering
\caption{\textbf{Coverage} on \textsc{DeformThings4D-Matching}}\label{tab:append:deform_cov}
\footnotesize
{\def\arraystretch{1.05}\tabcolsep=0.15em
\begin{tabular}{ccccccc}\toprule[0.8pt]
\multirow{2}{*}{methods} & \multicolumn{4}{c}{\textbf{\textit{near-isometric}}} & \textbf{\textit{partial}} & \textbf{\textit{non-iso}} \\ \cmidrule(lr){2-5}\cmidrule(lr){6-6}\cmidrule(lr){7-7}
 & \textsc{Zlorp} & \textsc{Drake} & \textsc{Mannequin} & \textsc{Ninja} & \textsc{Prisoner} & \textsc{Zombie}  \\ \midrule[0.8pt]
Init & $22$\% & $34$\%  & $35$\%  & $28$\%  & $8$\%  & $20$\%   \\ \midrule[0.5pt]
Ours w/ ARAP  & $28$\% & $37$\%  & $36$\%  & $34$\%  & $22$\%  & $29$\%  \\
Ours w/ nICP & $38$\% & $50$\%  & $53$\%  & $53$\%  & $20$\%  & $30$\%    \\
Ours w/ Shells & $61$\%  & $55$\%  & $56$\%  & $57$\%  & $39$\%  & $43$\%   \\\midrule[0.5pt]
ZO & $72$\%  & $70$\%  & $71$\%  & $71$\%  & $59$\%  & $60$\%   \\
DO & $68$\% & $66$\%  & $65$\%  & $66$\%  & $55$\%  & $55$\%    \\ \midrule[0.5pt]
Ours w/ D & $59$\% & $55$\%  & $54$\%  & $56$\%  & $37$\%  & $41$\%    \\
Ours w/ RHM & $64$\%  & $60$\%  & $60$\%  & $60$\%  & $45$\%  & $47$\%   \\ \bottomrule[0.8pt]
\end{tabular}}
\end{table}

\begin{table}[!t]
\centering
\caption{\textbf{Smoothness} via Conformal Distortion on \textsc{DeformThings4D-Matching}}
\label{tab:append:deform_smooth}
\footnotesize
{\def\arraystretch{1.05}\tabcolsep=0.15em
\begin{tabular}{ccccccc}\toprule[0.8pt]
\multirow{2}{*}{methods} & \multicolumn{4}{c}{\textbf{\textit{near-isometric}}} & \textbf{\textit{partial}} & \textbf{\textit{non-iso}} \\ \cmidrule(lr){2-5}\cmidrule(lr){6-6}\cmidrule(lr){7-7}
 & \textsc{Zlorp} & \textsc{Drake} & \textsc{Mannequin} & \textsc{Ninja} & \textsc{Prisoner} & \textsc{Zombie}  \\ \midrule[0.8pt]
Ours w/ ARAP & $2.33$ & $2.99$  & $2.21$  & $2.24$  & $3.02$  & $2.10$    \\
Ours w/ nICP  & $4.14$ & $5.15$  & $2.59$  & $2.90$  & $10.49$  & $4.58$   \\
Ours w/ Shells & $3.22$ & $4.68$  & $3.77$  & $4.56$  & $14.04$  & $7.13$    \\\midrule[0.5pt]
ZO & $3.05$ & $5.03$  & $2.51$  & $4.22$  & $24.80$  & $15.76$    \\
DO  & $3.23$ & $5.30$  & $3.77$  & $4.69$  & $21.10$  & $16.09$   \\ \midrule[0.5pt]
Ours w/ D & $2.85$ & $3.70$  & $2.81$  & $3.05$  & $9.89$  & $4.54$    \\
Ours w/ RHM   & $2.88$ & $3.92$  & $2.78$  & $3.07$  & $10.05$  & $4.72$   \\ \bottomrule[0.8pt]

\end{tabular}}
\end{table}

\section{Parameters}

In all experiments, we use the same set of parameters, where those of each smoothness energy were tuned independently. Parameters can also be found in the released implementation at \url{https://github.com/RobinMagnet/smoothFM}.

\paragraph{Spectral Energy.} For all experiments we weighted the spectral bijectivity term by $1$ and the coupling term by $10^{-1}$, as advocated in the Discrete Optimization implementation~\cite{ren2021discrete}.

\paragraph{Smoothness Energy.} Each smoothness energy required its own set of parameters. The Dirichlet energy was weighted by $1$ for all of them for consistency. In particular, for RHM energy, we used a coupling weight of $1$ and a bijectivity weight of $10^4$.   We used a coupling weight of $10^{-1}$ for ARAP, $10^{-2}$ for nICP and $10^{-3}$ for Shells.

\paragraph{Coupling.} We globally reweighted the smoothness energy by a parameter $\gamma$, gradually increasing from $10^{-1}$ to $1$ across iterations.

\section{Initialization}
\label{sec:appendix:Init}
For all datasets, we obtain initial dense correspondences by computing a $5\times5$ functional map using $5$ landmarks.

We chose this kind of initialization as standard shape descriptors like WKS~\cite{aubry2011wave} could not provide meaningful correspondences in the presence of high levels of non-isometry.

Indeed, Table~\ref{tab:append:tosca_wks} provides results using WKS descriptor as initialization for all methods. Note that the accuracy is unable to significantly go down from initialization. It thus becomes difficult to read into these results in a meaningful manner.

\begin{table}[!t]
\centering
\caption{Results on TOSCA-nonIso using WKS initialization.}
\label{tab:append:tosca_wks}
\footnotesize
{\def\arraystretch{1}\tabcolsep=0.8em
\begin{tabular}{ccccc}\toprule[0.8pt]
methods & \bfseries\itshape accuracy & \bfseries\itshape bijectivity & \bfseries\itshape smoothness & \bfseries\itshape coverage \\ \midrule[0.8pt]
Init & $56.56$  & $39.50$  & $93.24$ & $15.48$ \% \\\midrule[0.5pt]
Zo & $54.61$  & $43.23$  & $19.27$ & $52.48$ \% \\
DO & $53.65$  & $2.33$  & $16.47$ & $50.04$ \% \\\midrule[0.5pt]
Ours w/ D & $51.38$  & $22.30$  & $2.46$ & $16.72$ \% \\
Ours w/ RHM & $54.07$  & $4.18$  & $3.92$ & $35.29$ \% \\
\bottomrule[0.8pt]
\end{tabular}}
\end{table}

\section{Discrete Optimization}
\label{sec:appendix:Discrete Optimization}
The discrete optimization framework~\cite{ren2021discrete} proposes a large set of spectral energies, along which the conformal energy promoting functional maps associated to conformal pointwise correspondences. While this energy does help smoothness, we did not notice significant improvements regarding discontinuities in the correspondences.

\begin{figure}
    \centering
    \begin{overpic}[width=.15\textwidth]{
    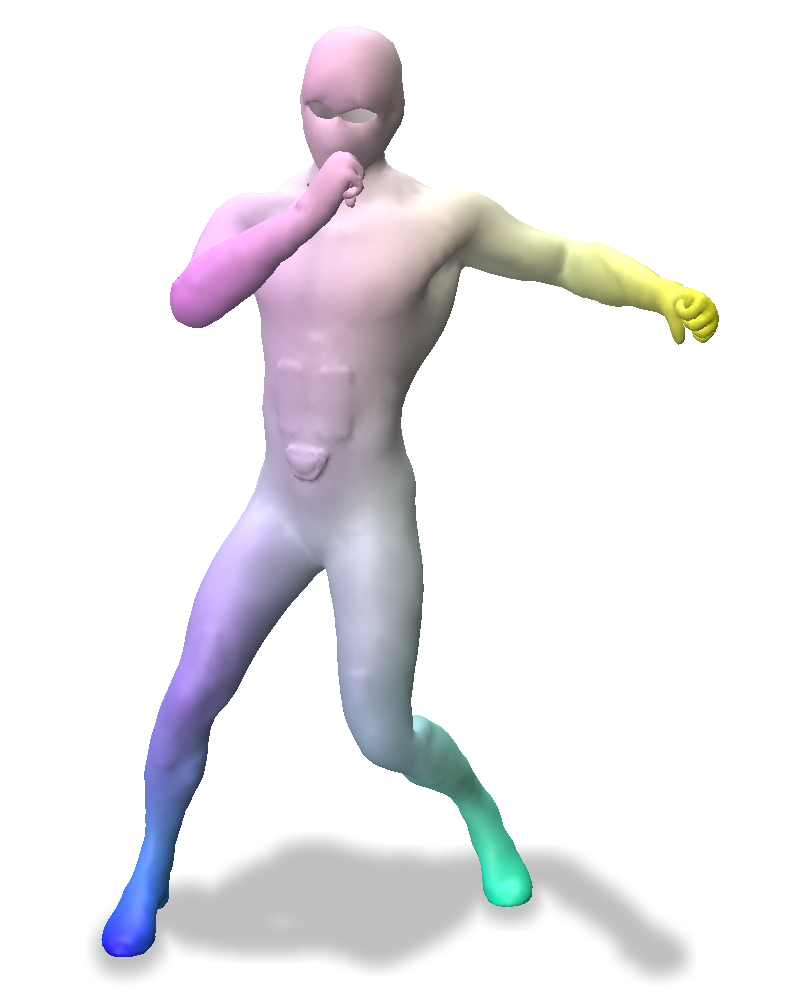}
    \put(22,102){\footnotesize Source}
    \end{overpic}
    \centering
    \begin{overpic}[width=.15\textwidth]{
    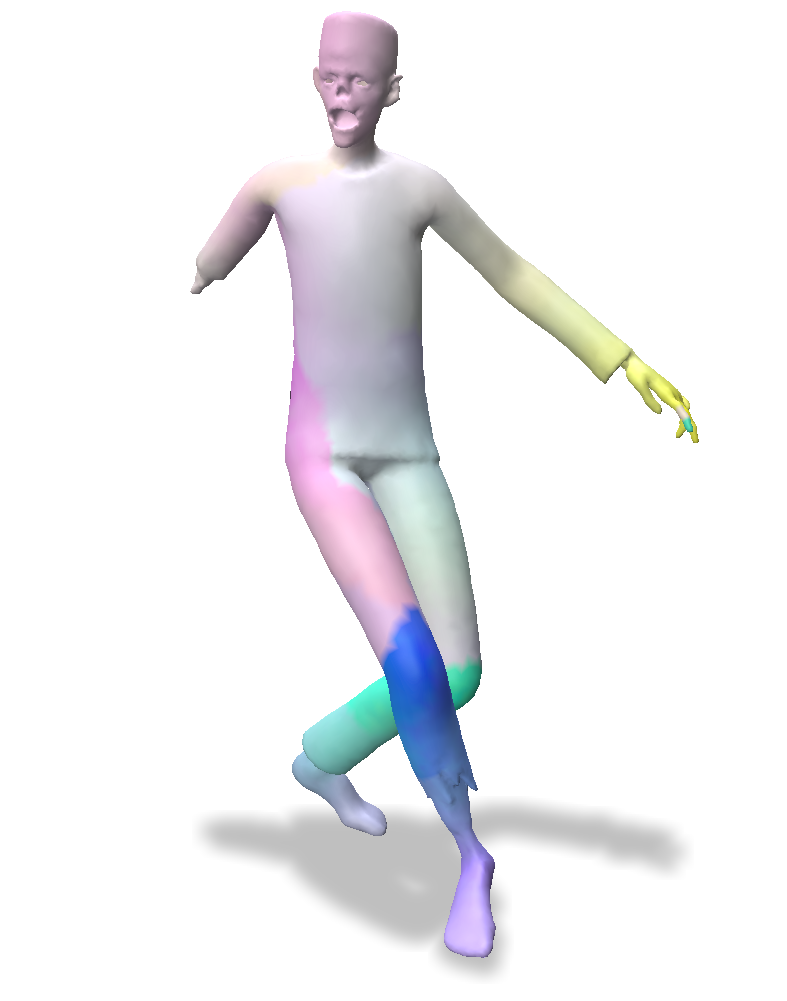}
    \put(25,102){\footnotesize DO}
    \end{overpic}
    \begin{overpic}[width=.15\textwidth]{
    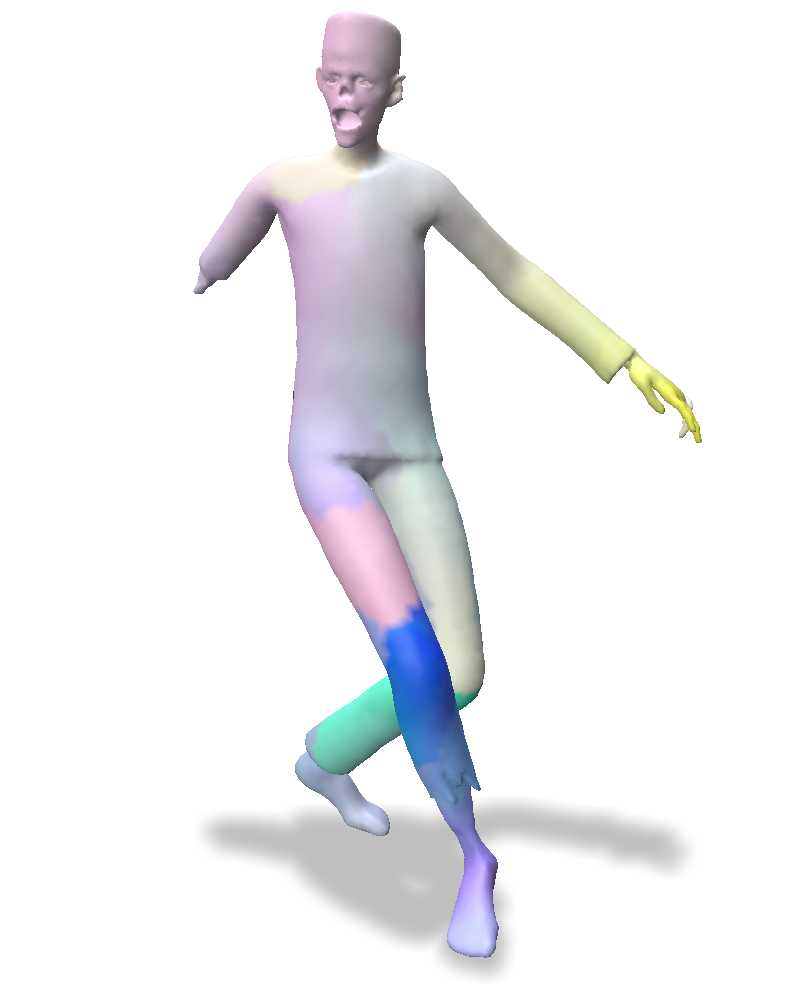}
    \put(22,102){\footnotesize DO + C}
    \end{overpic}
    \caption{Example of correspondences without (enter) and with (right) the conformal term of Discrete Optimization. While some parts are more smoother, the overall effect is marginal.}
    \label{fig:do_conf}
\end{figure}

On Figure~\ref{fig:do_conf}, we display an example of correspondences obtained by the standard Discrete Optimization (center) and by adding the conformal term (right). While some parts have been made smoother, the effect remain quite marginal.

In practice, this term can provide meaningful regularization in some cases but appears quite hard to tune to obtain a consistent effect.

\end{document}